\documentclass{article}
\usepackage{ijcai17}

\usepackage{times}

\usepackage{graphicx}
\usepackage{color}
\usepackage{pbox}
\usepackage{multirow}
\usepackage{caption}
\usepackage{subfig}
\usepackage{amsfonts}
\usepackage{amsmath}

\title{Efficient Dense Labeling of Human Activity Sequences from Wearables using Fully Convolutional Networks}
\author{Rui Yao$^{\dag \ddag}$, Guosheng Lin$^{\ddag}$, Qinfeng Shi$^{\ddag}$, Damith Ranasinghe$^{\ddag}$ \\
$^{\dag}$School of Computer Science and Technology, China University of Mining and Technology, China \\
$^{\ddag}$School of Computer Science, The University of Adelaide, Australia \\
ruiyao@cumt.edu.cn, \{guosheng.lin, qinfeng.shi, damith.ranasinghe\}@adelaide.edu.au}

\begin{document}

\maketitle

\begin{abstract}

Recognizing human activities in a sequence is a challenging area of research in ubiquitous computing.  Most approaches use a fixed size sliding window over consecutive samples to extract features---either handcrafted or learned features---and predict a single label for all samples in the window. Two key problems emanate from this approach: i) the samples in one window may not always share the same label. Consequently, using one label for all samples within a window inevitably lead to loss of information; ii) the testing phase is constrained by the window size selected during training while the best window size is difficult to tune in practice. 

We propose an efficient algorithm that can predict the label of each sample, which we call {\it dense labeling}, in a sequence of human activities of arbitrary length using a fully convolutional network. In particular, our approach overcomes the problems posed by the sliding window step. Additionally, our algorithm learns both the features and classifier automatically. We release a {\it new} daily activity dataset based on a wearable sensor with hospitalized patients. We conduct extensive experiments and demonstrate that our proposed approach is able to outperform the state-of-the-arts in terms of classification and label misalignment measures on three challenging datasets: Opportunity, Hand Gesture, and our new dataset. 

\end{abstract}

\vspace{-5mm}
\section{Introduction}
\label{sec:intro}

Recognizing human activities in sequences, known as human activity recognition (HAR), is a key research topic in human-computer interaction and human behaviour analysis in ubiquitous computing~\cite{bulling2014tutorial,Chavarriaga2013}. Among diverse human activities, some are more frequent or are of different durations than others. For example, an older person may spend prolonged periods in bed or sitting as opposed to ambulating. Consequently, class imbalance is an inherent nature of HAR problems. Time series signals in real applications are often collected from multiple sensors with very different characteristics. Thus it is very challenging to design good features and classifiers to recognize these activities both rapidly and accurately.    

A typical workflow of HAR methods for sequential data collected from wearable sensors contains the following steps: preprocessing, segmentation, feature extraction, and classification. Many works focus on handcrafting effective features such as signal-based feature~\cite{ravi2005activity}, transform-based feature~\cite{huynh2005analyzing}, and multilevel features~\cite{zhang2012motion}.  Another research avenue focus on using powerful learning algorithms to train classifiers (using existing features) such as Support Vector Machines (SVMs)~\cite{bulling2012multimodal}, boosting~\cite{blanke2009daily}, and temporal probabilistic graphical models such as Hidden Markov Model (HMM)~\cite{oliver2002layered}, conditional random field (CRF)~\cite{van2008accurate}, and semi-Markov models~\cite{shi2011human}.

Due to the success of deep learning~\cite{krizhevsky2012imagenet,simonyan2014very}, researchers have recently started to learn HAR features in an end-to-end fashion instead of handcrafting them. \cite{plotz2011feature} learned the layers of the autoencoder network with restricted Boltzmann machine for HAR. \cite{zeng2014convolutional,yang2015deep} proposed convolutional neural networks (CNNs) based approaches to automatically extract discriminative features for HAR. \cite{lane2015deepear} utilized CNNs to model audio data for ubiquitous computing (ubicomp) applications. \cite{hammerla2016deep,ordonez2016deep} investigated different types of deep neural networks for HAR with wearable sensors data. \cite{neverova2016learning} explored temporal deep neural networks for active biometric authentication. These deep learning based methods outperform other methods due to their ability of learning better features (than handcrafted ones).

However, regardless of using hand-crafted features or learned features, the input sensor data are always segmented into sections, typically using a sliding-window. It is hoped that each segment only contains a single activity. Given a sliding-window, one label is generated for all samples within the window. While widely used, this has several of issues. First, the sliding-window procedure creates the difficulty of defining the best window length, sampling stride and window labeling strategy; Second, the samples in one window may not always share the same ground truth label. People resort to intuitions and heuristics such as majority voting to force all samples to take one label. This inevitably loses original information, and causes label inconsistency---true labels or ground truth of some samples are not the label of the window and this can misguide the classifier; Third, for current sliding window based CNN methods, the window size for testing data must be the same as the window size used during the training. The imposes two issues. If users believe a different window size is better for the new testing data, they either have to stick to the old window size during the training (which they believe will be worse), or re-train the model from scratch. There are cumbersome ways to introduce some flexibility to the window size, but at the cost of speed and efficiency. Then the need to accumulate data over a fixed window can lead to intolerable prediction delays in real-time applications.

To manage these issues, we propose a method that can efficiently predict labels for each individual sample (which we call dense labeling) without any window based labeling procedure. Our contributions are three folds: i) We are the first to do dense labeling for HAR and avoid the label inconsistency problem caused by all sliding window approaches; ii) Our algorithm based on a fully convolutional network is much more efficient than CNN counter-parts, and can handle sequences of arbitrary length without window size restrictions; and iii) We release a new daily activity dataset collected from hospitalised older people which we believe will be beneficial to the community.  

\begin{figure}[t]
\centering
\includegraphics[width=.45\textwidth]{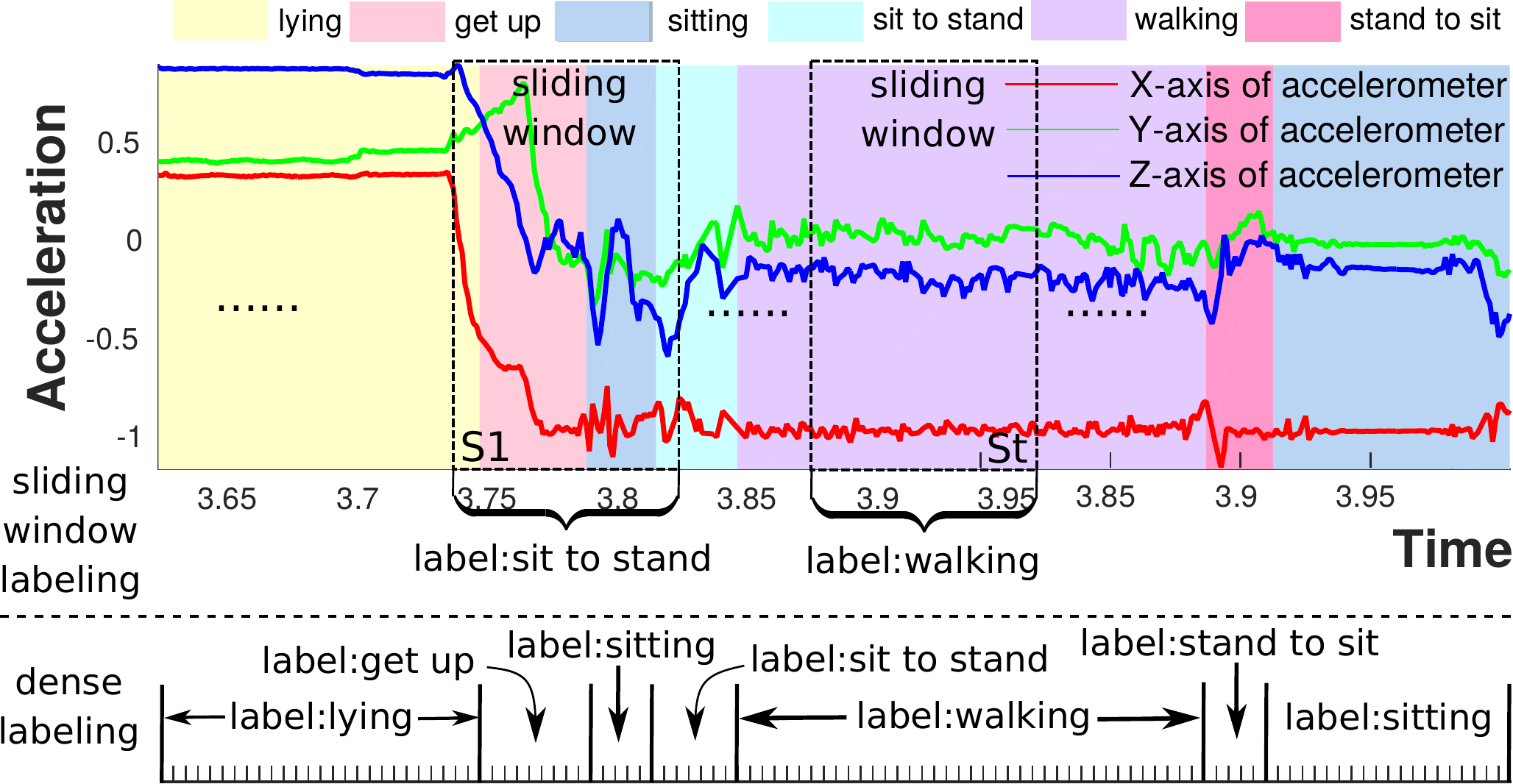}
\caption{Illustration of different labeling schemes. The top of the figure shows a segment of sensor data collected from a hospitalized patient where 
different classes of activity are shown in different colors. {\it S1} and {\it St} show two example sliding window instances and the predicted labels. The last row shows the predicted labels from dense labeling. The fixed size of the sliding  window {\it S1} will cause the loss of label information; it is hard to pre-segment windows dynamically by determining activity boundaries. We can avoid this problem with dense labeling by assigning a label to each sample.}
\label{fig:labeling}
\vspace{-3.9mm}
\end{figure}

\vspace{-3mm}
\section{Dense Sequence Labeling}
\label{sec:dense}

Human activity recognition can be viewed as a sequence labeling problem that estimates the class label of activity at each time step in long-term time sequence. The recent human activity recognition methods usually use a sliding window of fixed length to segment the data, and assign a class label to the window base on the label of the last time step (data point) in the window as shown in Fig.~\ref{fig:labeling}. However, segmenting a continuous sensor stream is a difficult task, and the exact boundaries of an activity are difficult to define~\cite{bulling2014tutorial}. As illustrated in Fig.~\ref{fig:labeling}, for example, the time series sensor data contains six-class activities. 
With a fixed sliding window size, one may annotate the class of sliding window $S1$ as \emph{sit to stand}. However, the window $S1$ contains four-class activities. Class \emph{lying, get up}, and \emph{sitting} are lost in window $S1$. 
Instead of making inaccurate segments that miss activity boundaries, we propose to train and predict the class of every sample densely and without window based labeling as shown in the bottom of Fig.~\ref{fig:labeling}.  Our method is efficient because  we are able to directly generate the dense labels of the sequence in one forward pass of our network without the need for subsampling. 

\begin{figure}[t]
\centering
\includegraphics[width=.45\textwidth]{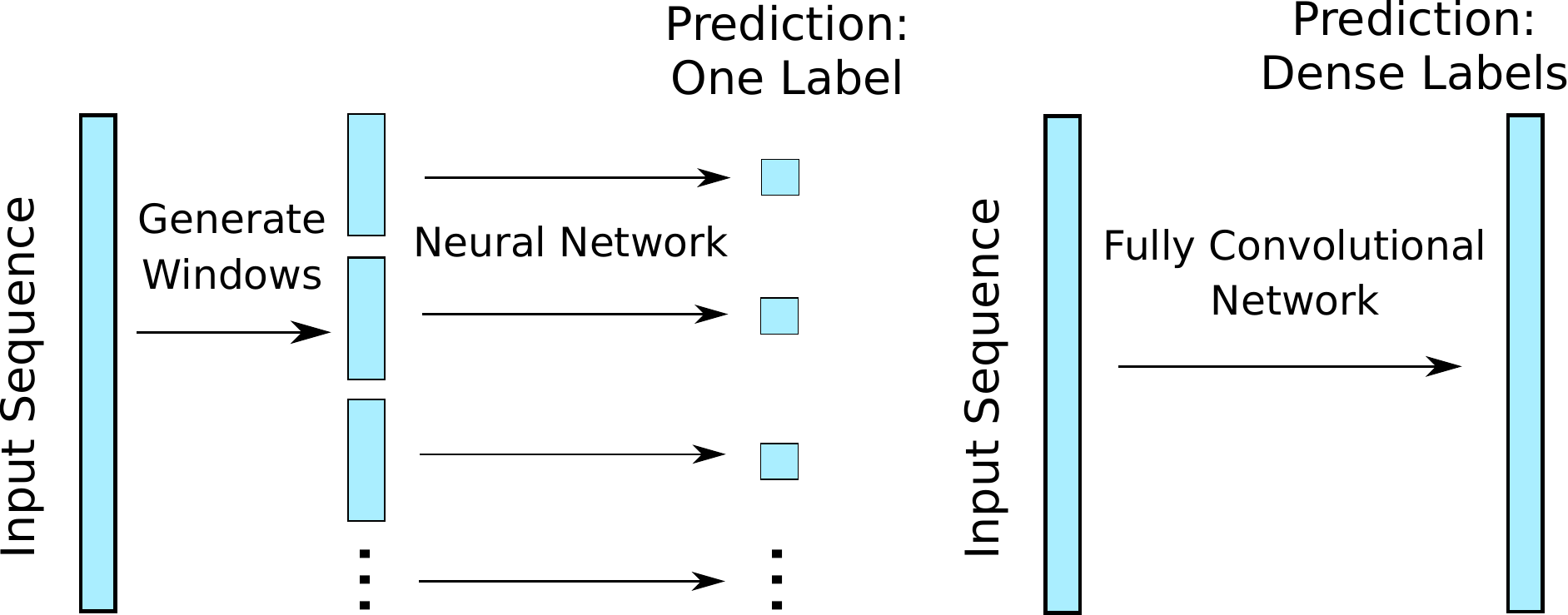}
\caption{Illustration of prediction procedure of conventional window based approach with CNN (left) and our dense approach with FCN (right). The window based CNN performs one forward pass for predicting one window, and the resulting labels are sub-sampled predictions. In contrast, we only need one forward pass of the network to generate predictions for the entire sequence  or a prediction for a single sample ({\it dense labeling}).}
\label{fig:cnn_fcn}
\vspace{-3.9mm}
\end{figure}


\begin{figure*}[t]
\centering
\includegraphics[width=1\textwidth]{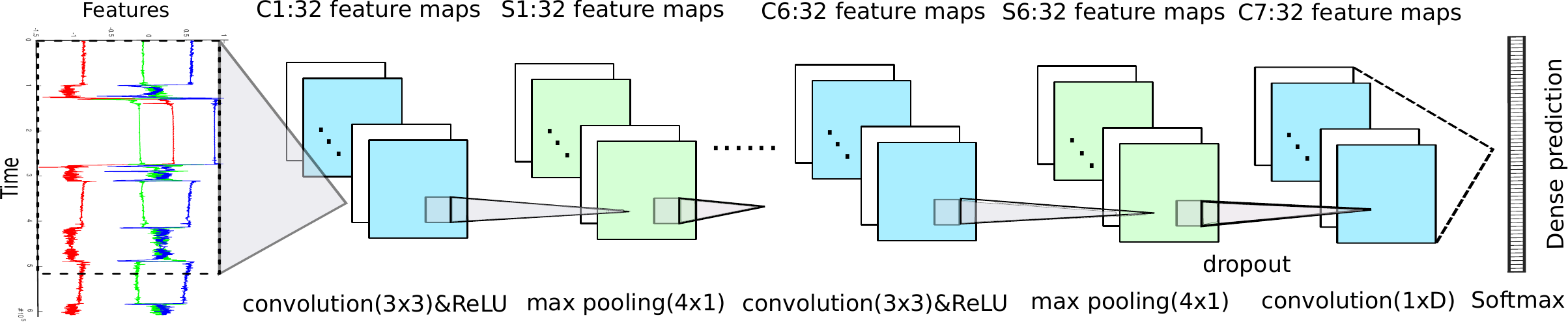}
\caption{Detailed view of the proposed fully convolutional network architecture. The input 2D time sequence is from the wearable sensors, where one dimension is time and the other is the attributes of all sensors. The proposed model repeats the layers convolution, ReLU, and max pooling six times, where the kernel size is reported in brackets. One dropout layer then performs zeroing operation, followed by the 7-th convolutional layer, where $D$ is the dimension of input data. The final softmax layer performs dense prediction. To ensure the output of each layer is the same length as the input sequence, we add padding in each convolutional and pooling layers.}
\label{fig:architecture}
\vspace{-3.9mm}
\end{figure*}

In the traditional method, generating windows for every time step, which we denote as setting the sampling stride to 1, will also achieve dense predictions of the sequence. However this straight forward and naive approach will result in a huge number of windows and the prediction process will became intractably slow. Specifically, in a conventional CNN based window classification approach, every window requires one forward pass of the CNN network, and it is clearly prohibitively expensive when handling a huge number of windows. This is also the reason that window based methods usually use a large sliding window stride to generate windows to subsample the labels of the original data. Here, we introduce the fully convolutional network (FCN) for this task. We are able to achieve dense and accurate predictions and, at the same time, only require low computational cost for the prediction process; these attributes make our approach an effective solution in practice. Fig.~\ref{fig:cnn_fcn} illustrates the different prediction procedure of conventional CNN and the FCN based network. We show the computational performance of the proposed method in Fig.~\ref{fig:time}.

\vspace{-3mm}
\section{Fully Convolutional Networks for Human Activity Recognition}
\label{sec:har}


The main component of the proposed method is a fully convolutional network, which is an extension of convolution neural networks~\cite{lecun1998gradient}. 
CNNs can be viewed as an enhancement of the standard Multi-Layer Perceptron (MLP), where the main difference is the addition of convolutional layers.
The basic components in CNN contain convolutional operation, pooling operation, and activation function. 
When CNNs are constructed by stacking these basic components layer by layer, a network that only contains the nonlinear filter
 is called \emph{Fully Convolutional Network} (FCN)~\cite{long2015fully}, which was originally introduced for the task of semantic segmentation. One advantage of FCN is that it can take input of arbitrary size and produce dense prediction with efficient inference and learning. To our best knowledge, we are the first to apply FCN into dense sequence labeling for HAR.

\vspace{-2mm}
\subsection{Network Architecture for Dense Labeling}
\label{subsec:arch}

Here we describe how to exploit FCN architecture for the problem of dense labeling of human activity sequences. The dense labeling estimates the likelihood of a particular activity at a certain time, and then predicts the class of the activity accordingly. The duration---length---of activities vary for different sequences and there are multiple activities in the input sequence, consequently, it is a naturally dense prediction task which requires label prediction for each sample
in the sequence. Therefore, it is reasonable to treat the recognition of human activity as a dense labeling problem, where the value at each location points out the confidence of each activity.

The proposed network architecture is shown in Fig.~\ref{fig:architecture}. The input to the network is time sequential data and its output is a dense confidence map of the input sequence. 
We denote the input of 
layer $l$ at location $(i,j)$ of filter channel $k$ as $\mathbf{x}_{i,j,k}^l$, where $1\leq i \leq D^l$ is the feature index, $1\leq j \leq T^l$ indicates the sample's location in total $T^l$ length sequence, and $1\leq k \leq N_f^l$ represents the index of filter. The input layer is connected by a convolution layer. For each filter $k$ of layer $l$, the output map is defined as a convolution with a kernel of size $f_i^l\times f_j^l$ followed by the addition of a constant ``bias'' term. To capture local temporal context
 we restrict each trainable filter with a small size. Let $\mathbf{W}_{i,j,k}^l$ be the components of the kernel and $\mathbf{b}_{k}^l$ the constant bias for channel $k$ of the layer $l$, respectively. Then the output of a specific location $(i,j)$ in channel $k$ of the layer $l$ is obtained by:
\begin{align}
\mathbf{z}_{i,j,k}^l = \sigma\big(\mathbf{b}_{k}^l + \sum_{\Delta i=-\bar{f_i^l}}^{\bar{f_i^l}}\sum_{\Delta j=\bar{f_j^l}}^{\bar{f_j^l}} &\sum_{k'=1}^{N_f^l} \mathbf{W}_{\Delta i + \bar{f_i^l}, \Delta j + \bar{f_j^l}, k'}^l \nonumber \\ &\mathbf{x}_{i+\Delta i, j+\Delta j, k'} \big),
\end{align}
where $\sigma$ is the activation function that introduces the nonlinearity into neural networks and allows it to learn more complex models. In this paper, we use Rectified Linear Units (ReLU)~\cite{nair2010rectified} as nonlinearity function in all activation layers. ReLU has been shown to work very well on deep convolutional networks for object recognition, and enforces small amounts of sparsity in neural networks. The ReLU activation function is defined as $\sigma_{ReLU}=\max(0,a)$. The convolution and activation layers are followed by a pooling layer. The purpose of pooling layer is to 
makes it robust to small variations in previously learned features, which is similar to a smoothing operation. The pooling strategy used in this work is maxpooling, which computes maximum value in each neighborhood at different positions. Recalling the architecture described in Fig.~\ref{fig:architecture}, we have kernel size $f_i^l=3$, $f_j^l=3$, and the number of filter $N_f^l=32$ in the first convolutional layer, and $f_i^l=1$, $f_j^l=4$, $N_f^l=32$ in the first pooling layer.

As illustrated in Fig.~\ref{fig:architecture}, we repeat those three layers (\emph{i.e.}, convolution, ReLU, and maxpooling), in sequence, six times with the same parameters. Then, the output is passed through a dropout layer that randomly set part of the inputs to zero with a specific probability. Finally, unlike the previous work using CNN architecture, we replace the fully connected layers with convolutional layers. There are $N$ filters in the final convolutional layer, where $N$ is the number of classes of human activities. The kernel size of each filter is $D^{l-1}\times 1$. The last layer of the proposed network is a 
dense softmax loss layer. Each unit of the output reflects a class membership probability: $p(z=c|\mathbf{x},\mathbf{W},\mathbf{b})=\frac{\exp(\mathbf{W}_c\mathbf{x}+\mathbf{b}_c)}{\sum_{c'}\exp (\mathbf{W}_{c'}\mathbf{x}+\mathbf{b}_{c'}) }$. Thus, we can obtain the posterior probability of the dense sequence classification results.

In contrast with the previous CNN methods for human activity recognition~\cite{yang2015deep,hammerla2016deep,zeng2014convolutional}, we make three important modifications to make the network architecture more suitable for the task of dense labelling of sequences for HAR: i) we replace the fully connected layers with the convolutional layers where the kernel size is $D\times 1$ and stride is $1\times 1$---$D$ is the dimension of input sequence; ii) for our convolution and pooling layers, we use padding to compensate for the kernel size and ensure the output is the same size as the input as shown in Fig.~\ref{fig:architecture}, although the number of filters may change; and iii) 
 we apply the dense softmax loss layer which densely impose the softmax classification loss for the predictions of all time steps. 
This is different from the existing methods which apply the conventional softmax loss layer for the prediction of a single window.



\subsection{Efficient Learning and Prediction}
\label{subsec:infer}

The replacement of convolutional layer is very efficient for processing large length time sequences for dense labeling  compared with applying sliding windows with the classical CNNs. In computer vision tasks, an FCN is able 
to classify the whole image at once by considering the summation of loss function of each individual loss at a specific location. However, in human activity recognition problems, we are not able to take the entire long-term time sequence as input to the network. It is computationally and practically prohibitive due to the infinite nature of a time series. In the proposed method, we separate a long-term time sequence into several overlapped subsequences to perform dense prediction. We define~\emph{subsequence} as a partition of length $T$ of contiguous samples from a whole sequence with starting position at $t$. By doing sliding step, we can extract a large amount of overlapped subsequences from a long-term time sequence. Notice that, the proposed \emph{subsequence} is not the same as sliding window, because we do dense prediction for each sample in a subsequence---\emph{i.e.}: one label for each sample---instead of predicting the sliding window as a whole---\emph{i.e.}: one label for all samples. Generally, the length of a subsequence can be far greater than the length of a window. We can define subsequences with sufficiently large length as shown in Fig.~\ref{fig:time}. 


The parameters of the proposed network are trained end-to-end by minimizing the negative log-likelihood function over the training set, summing over all samples in the subsequences:
\begin{align}
\ell (\mathbf{x},\mathbf{z};\mathbf{W,b})=\sum\nolimits_{j=1}^T \ell'(\mathbf{x},\mathbf{z}_j;\mathbf{W,b}),
\end{align}
where $\ell'(\mathbf{x},\mathbf{z}_j;\mathbf{W,b})=-\log \big(p(\mathbf{z}_j|\mathbf{x},\mathbf{W},\mathbf{b})\big)$ represents individual loss of $j$-th sample (data point corresponding to one time step) in a subsequence. The minimization of the loss function is performed by the Stochastic Gradient Descent (SGD) algorithm,  where we update the parameters after every subsequence. 
We compute the gradient over a mini-batch of the training subsequences. We randomly generate a number of subsequences to construct the mini-batch training data.

Once the parameters of the proposed FCN are learned from training data, we use the following two steps to get dense labeling of the whole time sequence. First, the dense confidence maps of each subsequence can be obtained by putting it into the proposed fully convolutional network. Second, because the two adjacent subsequences have overlapped samples, we have to find the overlapped samples and calculate the average probability of being a certain human activity. The final dense prediction (labeling) of time sequence is done by taking the argmax of the average probability of each samples. This prediction scheme is able to classify the whole subsequence at once, which is more efficient than using a sliding window scheme with a stride of one to perform dense labelling.

\begin{table*}[t]
\caption{Classification performance of the compared seven methods for locomotion and gesture task of \emph{Opportunity} dataset on \emph{all data, subject 1, subject 2}, and \emph{subject 3}. The highest performance is marked by boldface.}
\label{tab:opp}
\centering
\scalebox{.8}{
\begin{tabular}{c|c|c|c|c|c|c|c|c|c|c|c|c|c}
\hline
\multirow{2}{*}{Task} & \multirow{2}{*}{Method} & \multicolumn{3}{c|}{All Data} & \multicolumn{3}{c|}{Subject 1}  & \multicolumn{3}{c|}{Subject 2}  & \multicolumn{3}{c}{Subject 3}  \\
\cline{3-14}
& & $F_w$ & $NF_w$ & AC & $F_w$ & $NF_w$ & AC & $F_w$ & $NF_w$ & AC & $F_w$ & $NF_w$ & AC \\ \hline
\multirow{7}{*}{\rotatebox{90}{Locomotion}} & QDA & 76.0	&66.3	&67.5	&80.2	&71.3	&72.5	&75.9	&65.7	&67.7	&77.6	&68.0	&67.6	\\
 & KNN & 80.8	&80.1	&80.3	&85.5	&85.9	&85.9	&76.1	&77.9	&77.5	&78.2	&73.3	&72.7	\\
 & SVM & 80.7	&76.6	&77.2	&87.1	&86.3	&86.1	&78.5	&76.9	&77.2	&81.8	&75.2	&75.8	\\
 & Baseline & 83.6	&82.3	&82.4	&86.6	&85.9	&86.0	&80.3	&76.6	&76.9	&88.5	&82.4	&83.0	\\
 & CNN & 83.6	&82.6	&82.7	&87.7	&86.7	&86.8	&85.7	&86.0	&86.0	&88.5	&82.1	&82.4	\\
 & MLP & 82.6	&79.3	&79.7	&85.9	&84.5	&84.4	&80.8	&79.0	&79.1	&80.7	&73.4	&73.8	\\
 & Ours & \textbf{88.7}	&\textbf{86.9}	&\textbf{87.1}	&\textbf{91.9}	&\textbf{91.3}	&\textbf{91.2}	&\textbf{90.1}	&\textbf{89.8}	&\textbf{89.8}	&\textbf{89.0}	&\textbf{82.9}	&\textbf{83.2}	\\
 \hline
\multirow{7}{*}{\rotatebox{90}{Gesture}} &  QDA & 24.5	&58.8	&50.0	&19.3	&60.5	&52.8	&23.7	&69.1	&62.9	&27.1	&70.9	&67.2\\
 & KNN & 45.2	&85.2	&85.5	&45.6	&83.6	&84.2	&46.0	&85.5	&85.9	&49.6	&84.1	&85.7\\
 & SVM & 35.2	&82.7	&85.0	&36.1	&82.7	&85.8	&34.1	&82.6	&84.8	&25.3	&77.5	&82.3\\
 & Baseline & 42.9	&84.1	&84.4	&45.7	&84.9	&86.4	&13.2	&78.9	&83.2	&41.5	&82.1	&84.4\\
 & CNN & 48.8	&85.5	&85.7	&43.4	&83.8	&84.8	&42.1	&84.8	&86.2	&47.6	&84.0	&86.2\\
 & MLP & 39.7	&83.1	&83.7	&47.5	&84.9	&85.5	&42.9	&84.9	&85.8	&28.1	&78.3	&82.5\\
 & Ours & \textbf{59.6}	&\textbf{89.0}	&\textbf{89.9}	&\textbf{59.4}	&\textbf{89.3}	&\textbf{90.5}	&\textbf{55.0}	&\textbf{88.5}	&\textbf{89.7}	&\textbf{51.5}	&\textbf{85.6}	&\textbf{87.4}\\
 \hline
\end{tabular}
}
\end{table*}

\vspace{-2mm}
\section{Experiments}
\label{sec:exp}

We conduct extensive experiments to verify the performance of the proposed dense labeling method for HAR.
For FCN,
the learning rate for the whole network is set to $10^{-2}$ initially, decreased to $10^{-3}$ after 100 iterations, and  training is stopped at 150 iterations. The network weights are learned using the mini-batch (set to 10) SGD algorithm. The length of \emph{subsequence} is set to 60 for training, and 100 for testing. If there is enough memory, the length of subsequence can be set from 1 to the maximum length of whole test sequence. The adjacent subsequences has 50$\%$ overlap. For sliding window based methods, we set the window size to 24. Experiments were run on a machine with one GPU (NVIDIA GeForce GTX 770), all model are implemented using matlab\footnote{Our source code will be made available publicly.}. We test three non neural network classification techniques (\emph{i.e.}, Quadratic Discriminant Analysis (QDA), K-Nearest Neighbours (KNN, $k=3$)~\cite{Chavarriaga2013}, and Support Vector Machine (SVM)~\cite{chang2011libsvm}), and three neural network methods (baseline convolution network--Baseline, convolutional neural network based method by~\cite{yang2015deep}--CNN, and three-layer Multi-Layer Perceptron--MLP). The baseline convolutional network contains four convolution-ReLU-max pooling layers, and follows one fully connected layer and softmax layer. The Baseline, CNN and MPL use a traditional sliding window to generate classifications.  The learning rate and iteration parameters of Baseline are set to be the same as ours.

\vspace{-2mm}
\subsection{Dataset}
\label{subsec:dataset}

We evaluate the proposed approach on two public benchmark datasets and a new dataset released by our group.

{\bf Opportunity dataset.} The \emph{Opportunity} activity recognition dataset~\cite{5573462} comprises a set of daily activities collected in a sensor-rich environment. In this paper, we address two kinds of recognition problems: \emph{locomotion} and \emph{gestures}, and use the same subset employed in the \emph{Opportunity} challenge~\cite{Chavarriaga2013,ordonez2016deep} to train and test our models. We evaluate the performance on \emph{all data, subject 1, subject 2} and \emph{subject 3} of the dataset. We use the same training and testing set as in~\cite{Chavarriaga2013}.
 Locomotion includes 5 classes with \emph{null}-class, and gestures contains 18 classes with \emph{null}-class. Each sample is comprised of 113 real valued attributes, which are scaled to $[0,1]$ for training and prediction; the same data pre-processing method will be applied to the following two datasets.

{\bf Hand Gesture dataset.} The \emph{Hand Gesture} dataset \cite{bulling2014tutorial} records arm movements of two people (subject 1 and 2) performing a continuous sequence of different types gestures of daily living. The dataset has 12 classes with \emph{null} class, and the \emph{null} class refers to periods with no specific activity. The used data includes three three-axis accelerometers and three two-axis gyroscopes recoding at a sampling rate of 32~Hz. Each sample has 15 real valued sensor attributes in total. For training, testing and validation, we use the same approach as that by \cite{yang2015deep} to generate the data.

\begin{table}[t]
\caption{The classification performance of~\emph{Hand Gesture} dataset. The highest performance is marked by boldface.}
\label{tab:hand}
\centering
\scalebox{.8}{
\begin{tabular}{c|c|c|c|c|c|c}
\hline
\multirow{2}{*}{Method} & \multicolumn{3}{c|}{Subject 1} & \multicolumn{3}{c}{Subject 2}   \\
\cline{2-7}
 & $F_w$ & $NF_w$ & AC & $F_w$ & $NF_w$ & AC \\ \hline
QDA &  	 	50.2	 	&46.0	&46.7	&67.2	&63.9	&63.1\\
KNN &  	 	85.3	 	&84.5	&84.4	&85.1	&82.9	&82.7\\
SVM &  	 	53.6	 	&55.6	&57.5	&38.0	&39.9	&45.5\\
Baseline &  	 	77.4	 	&73.0	&72.2	&80.2	&77.4	&77.0\\
CNN &  73.7	 	&69.2	&68.1	&78.9	&76.2	&75.6\\
MLP &   	79.7	 	&80.1	&80.1	&77.1	&75.9	&75.5\\
Ours &  	\textbf{89.3}	 	&\textbf{88.2}	&\textbf{88.1}	&\textbf{88.3}	&\textbf{86.1}	&\textbf{85.8}\\
\hline
\end{tabular}}
\end{table}

{\bf Hospital dataset.} To verify the performance of the proposed method, we collect our own HAR dataset using an inertial sensor with one accelerometer and one gyroscope. Each sample has 5 real valued attributes in total. We recorded 8 activities from 12 volunteer hospitalised older patients. During the trial, participants were informed only of the types of activities they were required to perform and the inertial sensor was worn loosely over their garments above the sternum area to monitor the activities, which include: \emph{lying, stand up (sit to stand), sitting, walking, lie down (get into bed), sit down (stand to sit), get up,} and \emph{standing}. 
All data is recorded at a frequency of approximately 10~Hz. For all the compared methods, we set the data of the first eight volunteers as training data, the data from the following three volunteers are used for testing, and the last one for validation. This dataset will be publicly available at our project website. 

\subsection{Evaluation Protocol}
\label{subsec:measures}

Two kinds of evaluation protocols are employed in this paper: classification and misalignment measures.
\vspace{-5.3mm}
\paragraph {Classification measures:} We use three widely used metrics: weighted F-measure ($F_w$), wighted F-measure with \emph{null} class ($NF_w$), and accuracy (AC). Due to class imbalance in the three datasets, we calculate the F-measure  weighted according to their sample proportion: $F_w = \sum\nolimits_i 2\cdot w_i\cdot (p_i\cdot r_i) / (p_i + r_i)$
where $i$ is the class index and $w_i=n_i/N$ with $n_i$ the number of samples of the $i$-th class, $N$ the total number of samples and $p_i$ denotes precision while $r_i$ represents recall. Here $F_w$ measures  
performance without \emph{null} class.

\begin{table}[t]
\caption{The classification performance of for \emph{Hospital} dataset. The highest performance is marked by boldface.}
\label{tab:ade}
\centering
\scalebox{0.8}{
\begin{tabular}{c|c|c|c|c|c|c|c}
\hline 
 & QDA & KNN & SVM & Baseline & CNN & MLP & Ours \\ \hline
$F_w$ & 64.8 & 79.9 & 76.2 & 81.9 & 81.4 & 78.4 &  \textbf{88.2}\\ 
AC &  70.1 & 86.6 & 84.8 & 88.3 & 87.0 & 85.3 & \textbf{91.2}\\
\hline
\end{tabular}
}
\end{table}

\begin{figure*}[t]
\centering
\subfloat{\label{fig:legend}\includegraphics[width=.7\textwidth]{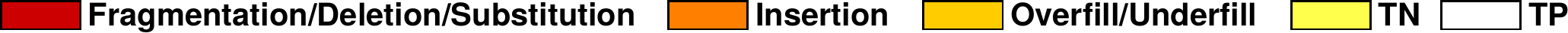}} \vspace{-1mm}
\addtocounter{subfigure}{-1}
\subfloat[Opportunity Dataset]{\label{fig:opp}\includegraphics[width=.41\textwidth]{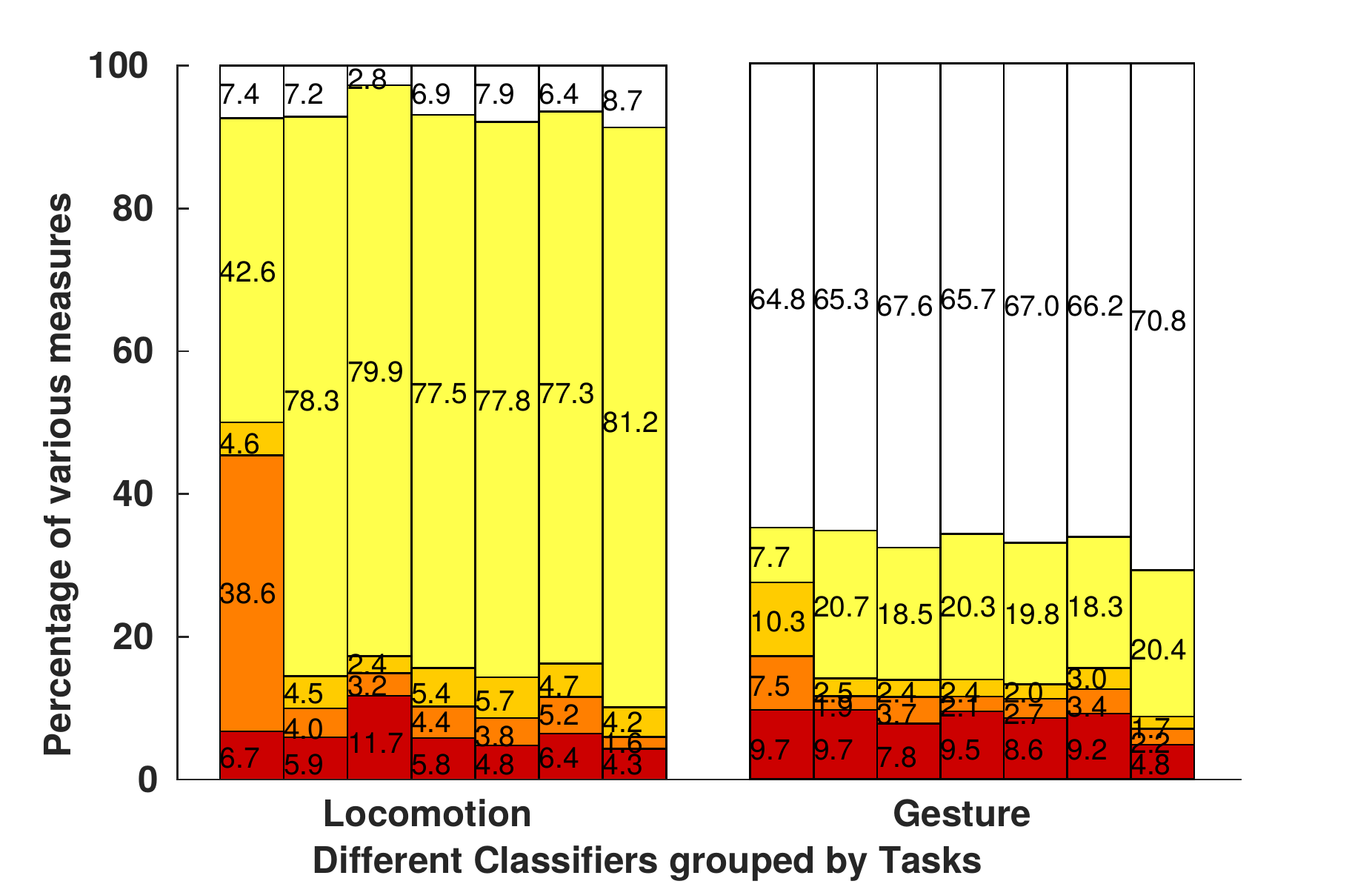}} \hspace{-5mm}
\subfloat[Hand Gesture Dataset]{\label{fig:hand}\includegraphics[width=.41\textwidth]{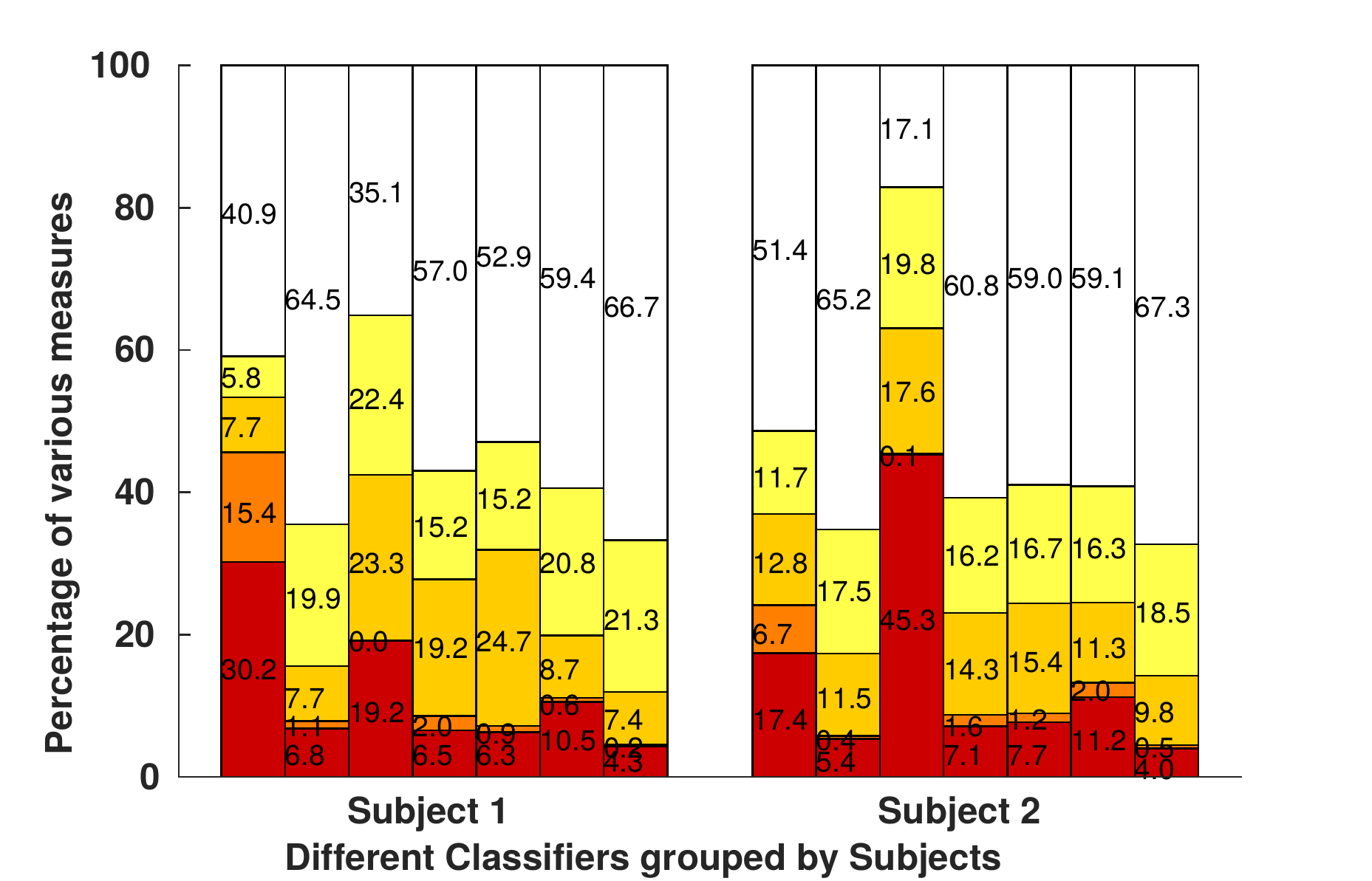}} 
\subfloat[Ours Dataset]{\label{fig:ade}\includegraphics[width=.202\textwidth]{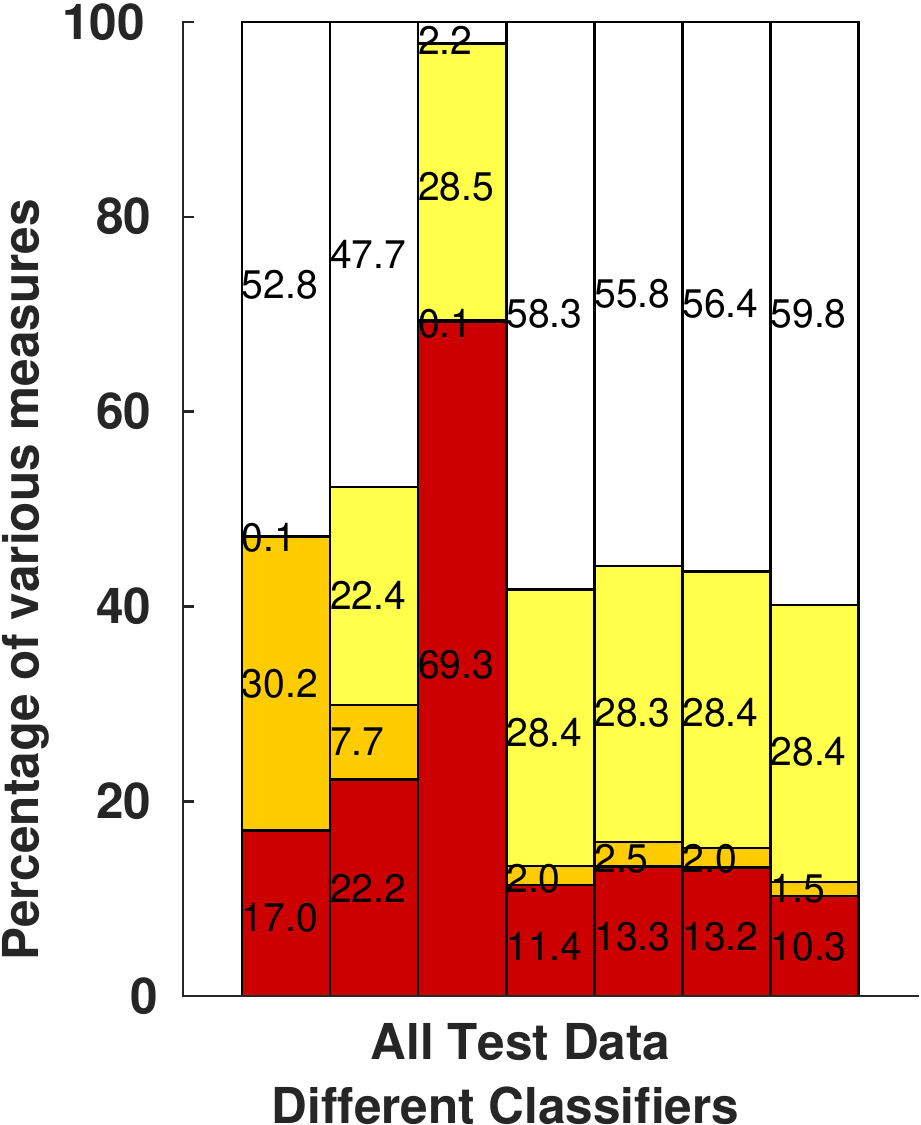}}
\caption{Performance evaluation on three datasets for different tasks using the measures proposed by~\protect\cite{ward2011performance} . Each group of seven columns denotes the accuracy of QDA, KNN, SVM, Baseline, CNN, MLP, and Ours. The digital number on each column shows the exact value of the corresponding measures. Overall our method performs the best.}
\label{fig:misalign}
\end{figure*}

\vspace{-5.mm}
\paragraph{Misalignment measures:} The classification measures may be misleading when recognizing actions from continuously recorded data. Therefore, we explicitly report different types of errors: \emph{True Positive--TP, True Negative--TN, Overfill/Underfill, Insertion}, and \emph{Fragmentation/Deletion/Substitution}~\cite{ward2011performance} to measure predicted label misalignments with the ground truth. 
Errors when the start or stop time of a predicted label is earlier or later than the actual time is an \emph{Overfill} while  a predicted label is later or earlier than the actual is an \emph{Underfill} measure.
Predicting an action when there is \emph{null} activity is an \emph{Insertion}.
\emph{Fragmentation} denotes the errors of predicting a \emph{null} class in between an uninterrupted activity class. \emph{Deletion} represents the errors of assigning a \emph{null} label when there is an activity. \emph{Substitution} measures the errors when an activity is misclassified as a different class other than \emph{null}.

\vspace{-2mm}
\subsection{Results}
\label{subsec:results}

Comparing the performance of different methods across different studies is difficult for many reasons such as the differing types of experimental protocols and feature used. To make a fair comparison, for QDA, KNN, and SVM methods, we used a sliding window with a fixed step size for extracting the mean value of the sensor on each window as features. The obtained results 
are quite similar to the benchmark results on the Opportunity dataset~\cite{bulling2014tutorial}.

\begin{figure}[t]
\centering
\includegraphics[width=.22\textwidth]{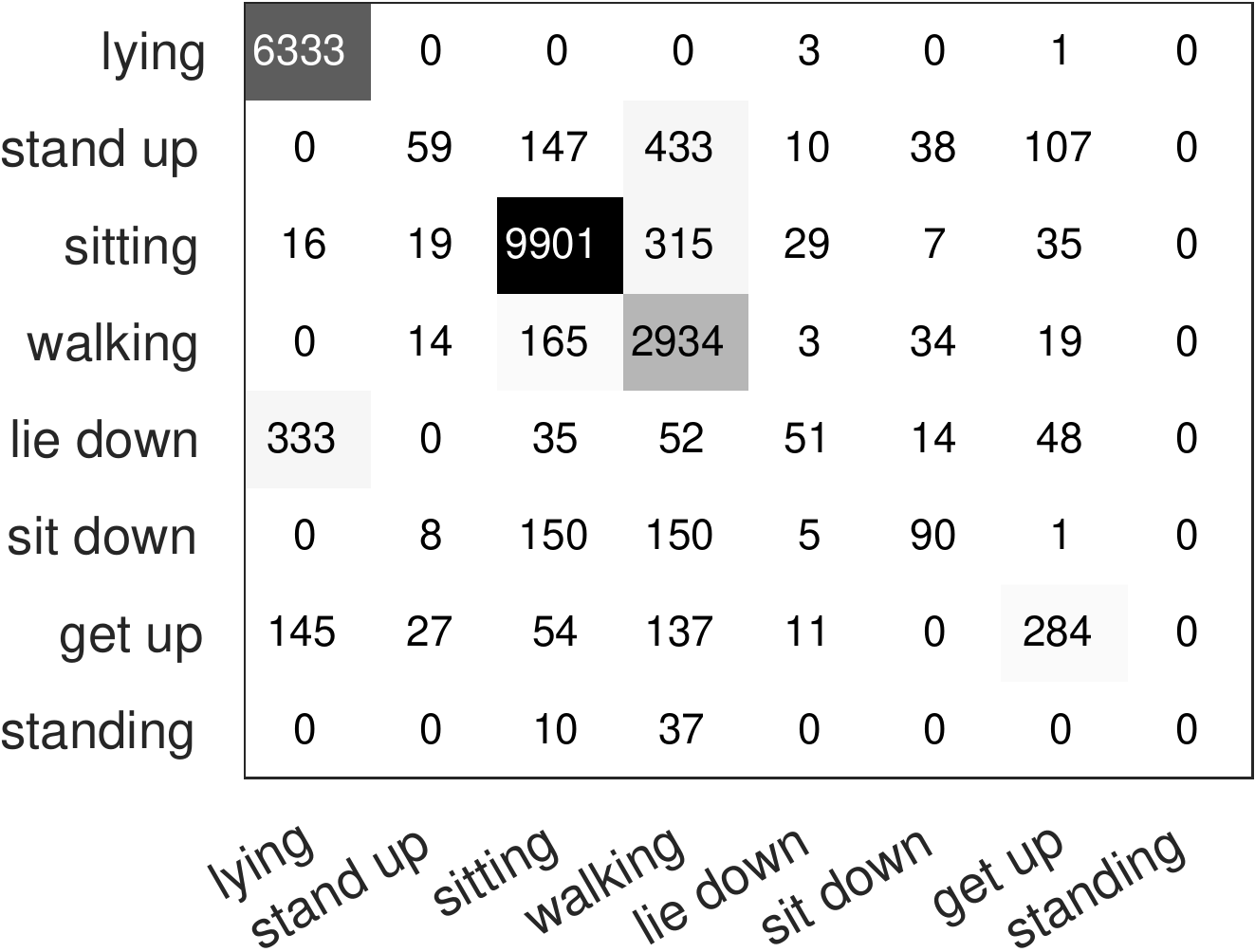} \hspace{1mm}
\includegraphics[width=.22\textwidth]{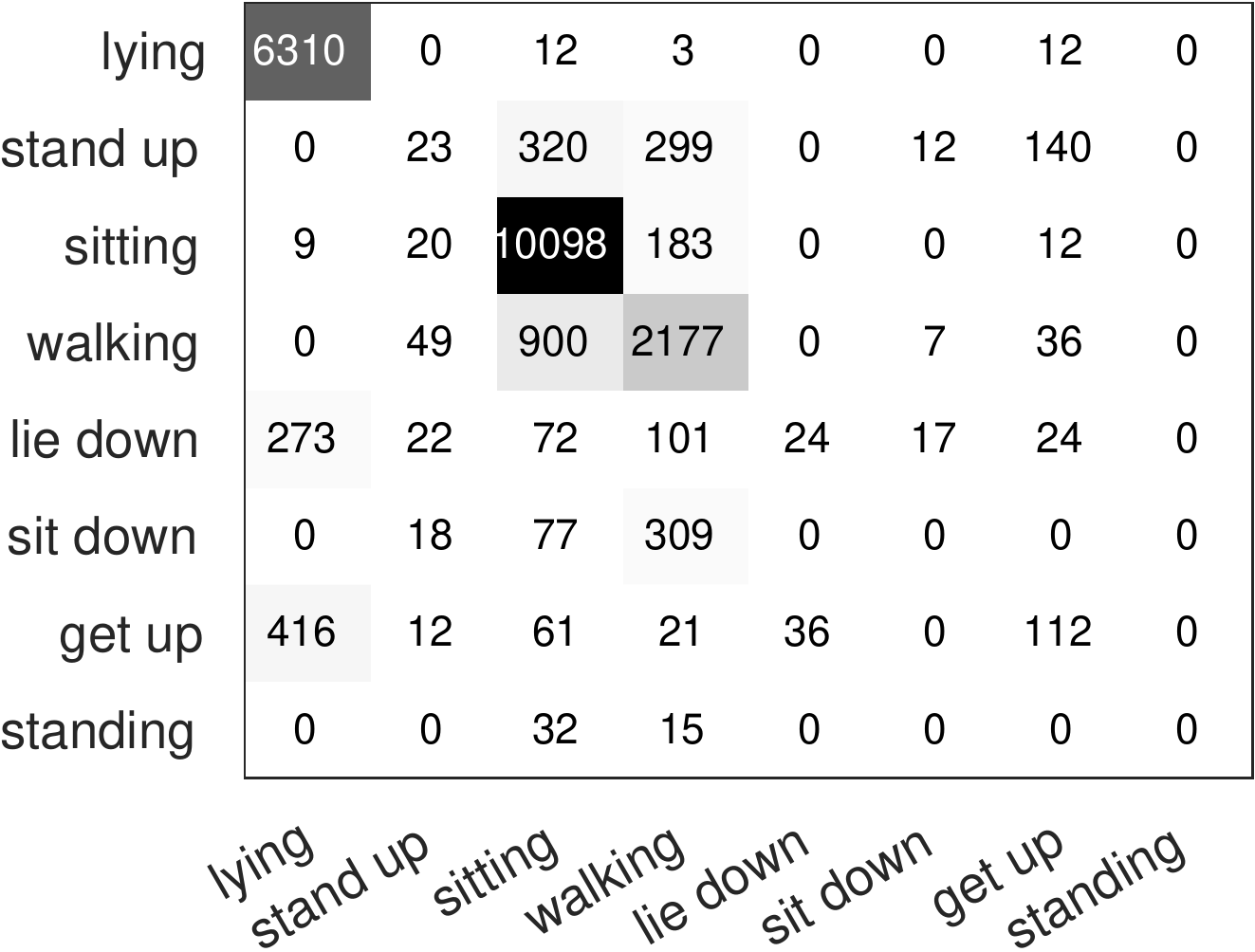}
\caption{Confusion matrix for \emph{Hospital} dataset using the proposed method (left) and CNN method (right). Our method is more robust against activity transitions (\emph{i.e.}: \emph{stand up, lie down, sit down}, and \emph{get up}).}
\vspace{-5mm}
\label{fig:conmat}
\end{figure}

We report the classification performance on the three datasets in Tab.~\ref{tab:opp}, Tab.~\ref{tab:hand} and Tab.~\ref{tab:ade}, respectively. In general, CNN-based methods perform better on almost all dataset than non-CNN methods and our approach, denoted {\it Ours}, performs better than all CNN-based methods.
Specially, class imbalance leads to overesting the performance of gesture recognition. {\it Ours} obtains $59.0\%$ with wighted F-measure on \emph{all} Opportunity data, which is $11\%$ higher than CNN methods proposed by~\cite{yang2015deep}. A similar conclusion can be drawn in \emph{subject 1} and \emph{subject 2} for gesture recognition. Those results demonstrate that the proposed dense labeling method is more robust with respect to the class imbalance problem. Note that, we do not report $NF_w$ in Tab.~\ref{tab:ade} because \emph{null} class is not included in our hospital dataset. Although the proposed approach obtains slight higher accuracy than the baseline method on the new dataset, the $F_w$ value of {\it Ours} in Tab.~\ref{tab:ade} is significantly higher than other methods. We advocate that the weighted F-measure can reflect the real performance of unbalanced datasets. Fig.~\ref{fig:conmat} shows the confusion matrix on \emph{Hospital} dataset generated by our method and CNN method proposed in \cite{yang2015deep}. 

\begin{figure}[t]
\vspace{-5mm}
\centering
\raisebox{3mm}{\includegraphics[width=.2\textwidth]{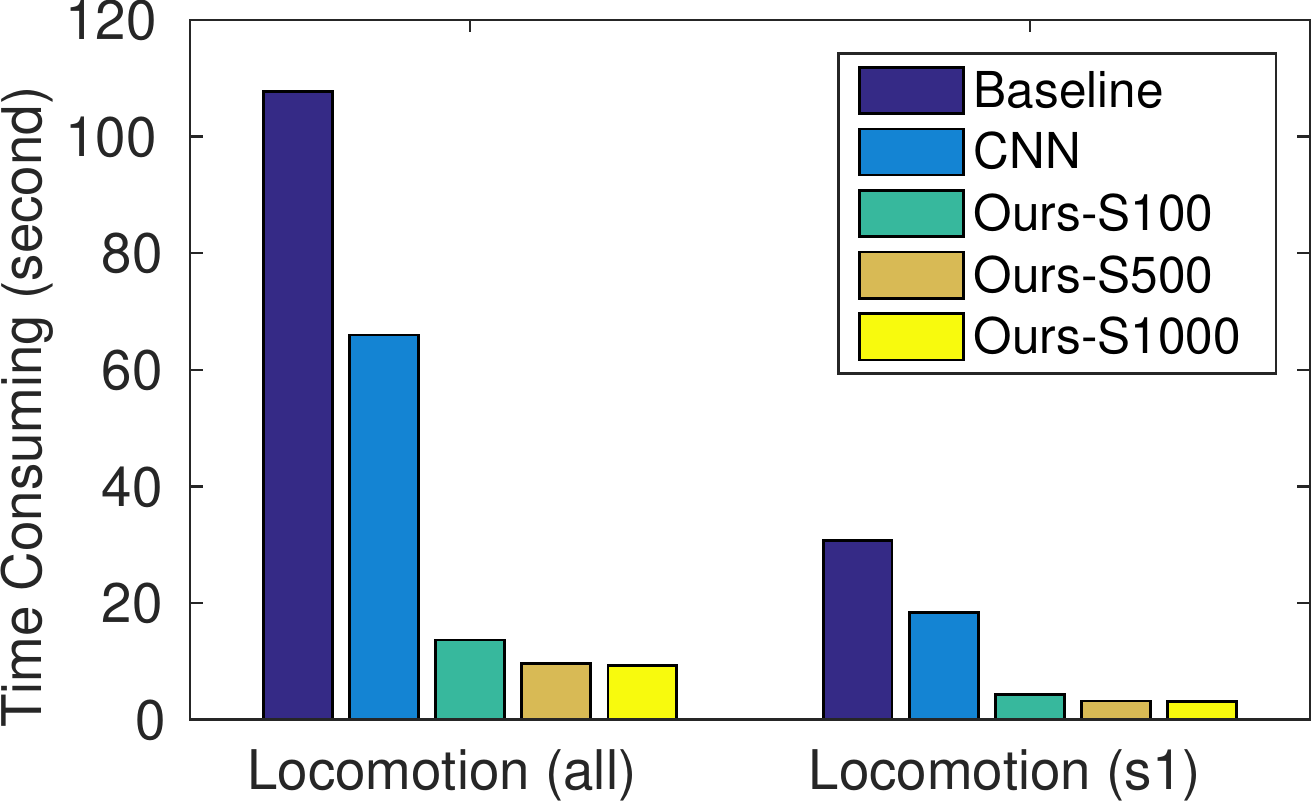} \hspace{5mm}}
\includegraphics[width=.22\textwidth]{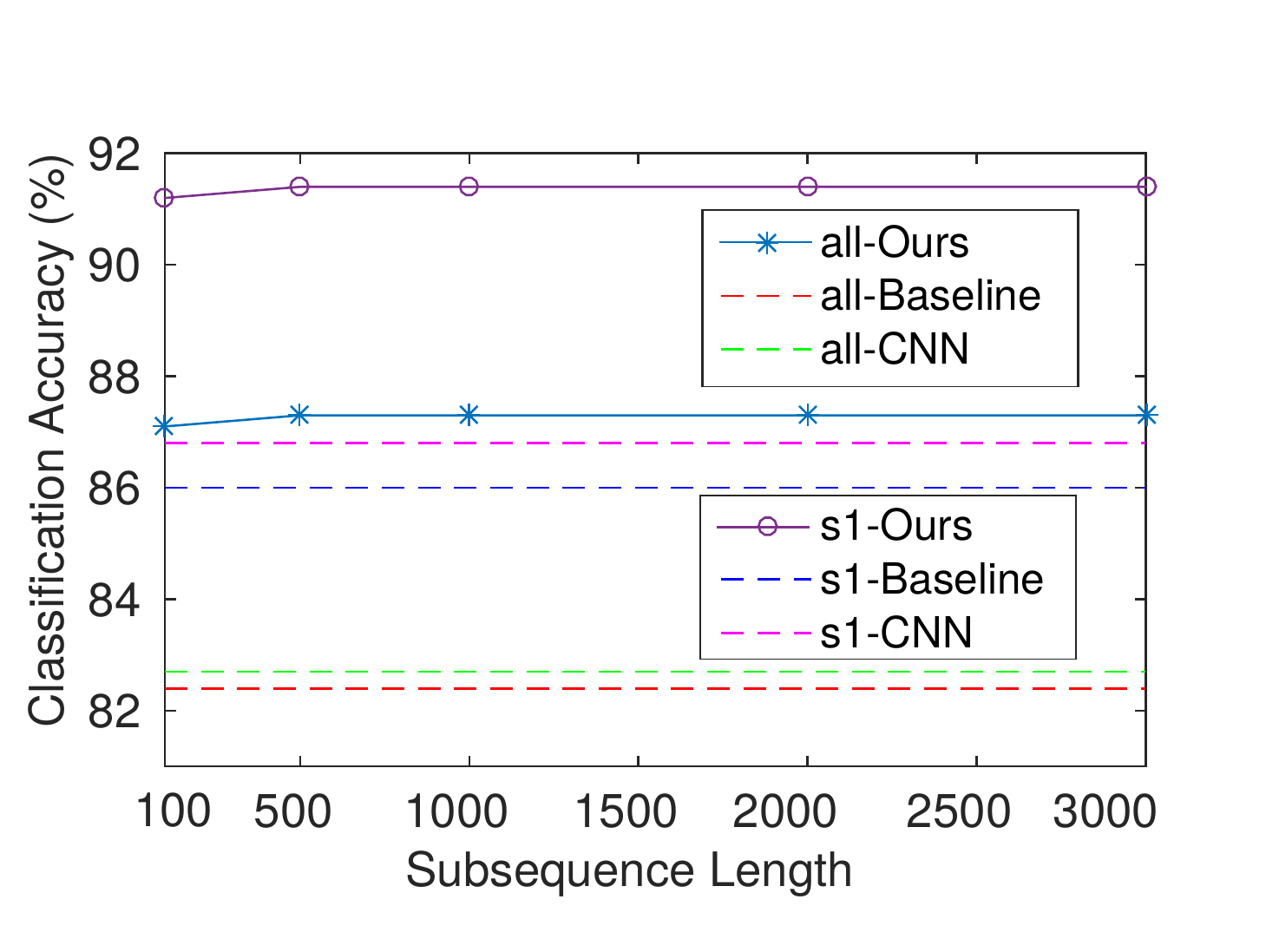}
\vspace{-3mm}
\caption{Testing time and accuracy of the compared convolutional networks in CPU-seconds. We can see the testing time (left) and classification accuracy (right) for task \emph{locomotion} on \emph{all} and \emph{subject 1 (s1)} of \emph{Opportunity} dataset. \emph{Ours-S100} denotes the length of the \emph{subsequence} used for testing our method as being 100. We can see that the performance of our method is insensitive to the length of the subsequence. Unlike traditional CNN based methods that require fixing the sliding window size for testing and training to be identical, a subsequence of our method can be of arbitrary length.}
\vspace{-3mm}
\label{fig:time}
\end{figure}

Fig.~\ref{fig:misalign} shows the misalignment performance on the three datasets. For clarity, we only report the overall performance of all Opportunity data in Fig.~\ref{fig:misalign}(a), and combine several measures to one term in contrast with the benchmark in~\cite{bulling2014tutorial}. It can be seen that {\it Ours} achieves the best performance among the compared methods. This demonstrates that the dense labeling method more accurately finds the starting and ending point of activities in the temporarily dependent sequence data. Because we do not contain a \emph{null} class in the Hospital dataset, \emph{Fragmentation/Deletion/Substitution} term in Fig.~\ref{fig:misalign} (c) only represents the error of \emph{Substitution}.


We report the computational performance of three convolutional neural networks on \emph{Opportunity} datasets in Fig.~\ref{fig:time}. Our approach only consumes one-fifth and one-tenth of the time CNN~\cite{yang2015deep} and Baseline approach consumed. There are 118,750 and 33,273 samples with 113-D features in \emph{all} and \emph{subject 1} of \emph{Opportunity} dataset. 
Note the maximum length of subsequence, 3000, is limited by the machine used for experiments (4GB memory, single GPU).  


\vspace{-3.9mm}
\section{Conclusion}
\label{sec:conclusion}

In this paper, we have presented a new, efficient dense labeling approach for human activity recognition using a fully convolutional network. Unlike existing methods, the proposed approach does not rely on a heuristics sliding-window step for producing ambiguous labels of the window---often a source of error during training and prediction. Our approach uses a fully convolutional network framework based on the well-studied deep convolutional neural network theory and therefore making it easy to train and infer the dense labels of sequences as well as benefit from generalization and robustness to extract feature end-to-end. We also release a new dataset publicly for human activity recognition. The promising results demonstrate the effectiveness and efficiency of the proposed dense labeling approach. 

\small{
\bibliographystyle{named}
\bibliography{ijcai17}
}

\end{document}